\documentclass[letterpaper, 10 pt, conference]{ieeeconf}  

\IEEEoverridecommandlockouts                              

\usepackage{kotex}
\usepackage{amsfonts}
\usepackage{amsmath}
\usepackage{textcomp}
\usepackage{multirow}
\usepackage{graphicx}
\usepackage{lipsum}
\usepackage{longtable}
\usepackage{booktabs}
\usepackage{dblfloatfix}
\usepackage{xcolor} 
\usepackage{hyperref}
\usepackage{xfrac}    
\usepackage{algpseudocode}
\usepackage{lipsum} 
\usepackage{wrapfig}
\usepackage{multicol}
\usepackage[ruled,vlined]{algorithm2e}

\usepackage{tabularx} 

\title{\LARGE \bf
Fast Global Localization on Neural Radiance Field
}

\author{Mangyu Kong$^{1}$, Jaewon Lee$^{1}$, Seongwon Lee$^{2,\dagger}$, and Euntai Kim$^{1,\dagger}$%
\thanks{$\dagger$ Corresponding authors.}%
\thanks{$^{1}$M.\,Kong, J.\,Lee, and E.\,Kim are with the School of Electrical and Electronic Engineering, Yonsei University, Seoul 03722, South Korea, {\tt\small \{mangyu0929,leejaewon,etkim\}@yonsei.ac.kr}.}%
\thanks{$^{2}$S.\,Lee is with the School of Electrical Engineering, Kookmin University, Seoul 02707, South Korea, {\tt\small sungonce@kookmin.ac.kr}.}%
}

\begin{document}

\maketitle
\thispagestyle{empty}
\pagestyle{empty}

\begin{abstract}

Neural Radiance Fields (NeRF) presented a novel way to represent scenes, allowing for high-quality 3D reconstruction from 2D images. Following its remarkable achievements, global localization within NeRF maps is an essential task for enabling a wide range of applications. Recently, Loc-NeRF demonstrated a localization approach that combines traditional Monte Carlo Localization with NeRF, showing promising results for using NeRF as an environment map. However, despite its advancements, Loc-NeRF encounters the challenge of a time-intensive ray rendering process, which can be a significant limitation in practical applications. To address this issue, we introduce Fast Loc-NeRF,  which enhances efficiency and accuracy in NeRF map-based global localization. We propose a particle rejection weighting strategy that estimates the uncertainty of particles by leveraging NeRF’s inherent characteristics and incorporates them into the particle weighting process to reject abnormal particles. Additionally, Fast Loc-NeRF employs a coarse-to-fine approach, matching rendered pixels and observed images across multiple resolutions from low to high. As a result, it speeds up the costly particle update process while enhancing precise localization results. Our Fast Loc-NeRF establishes new state-of-the-art localization performance on several benchmarks, demonstrating both its accuracy and efficiency. The code is available at \href{https://github.com/kmk97/Fast-Loc-NeRF}{this url}.

\end{abstract}

\section{INTRODUCTION}

Global localization is one of the most important challenges in the field of mobile robotics. The detailed algorithm used for global localization is inherently tied to the type of map employed. Different mapping techniques require tailored localization algorithms to determine a robot's position effectively without knowledge about the initial pose. In the early days of mobile robotics, some relatively simple yet efficient maps were commonly used, for example, feature-based maps such as ORB maps \cite{mur2017orb,campos2021orb} or 2D occupancy grid maps~\cite{grisetti2007improved}, but there is a rapid shift towards utilizing a photo-realistic volumetric 3D environment map. Neural Radiance Fields (NeRF)~\cite{mildenhall2021nerf} is certainly the most typical example of a realistic volumetric 3D map. The reason for this is that this type of 3D realistic volumetric map can be utilized not only in mobile robotics but also in immersive technologies such as AR/VR/MR. Therefore, global localization under the NeRF map emerges as one of the most urgent topics that should be considered in mobile robotics. 

\begin{figure}[htbp]
\centering
\includegraphics[width=\columnwidth]{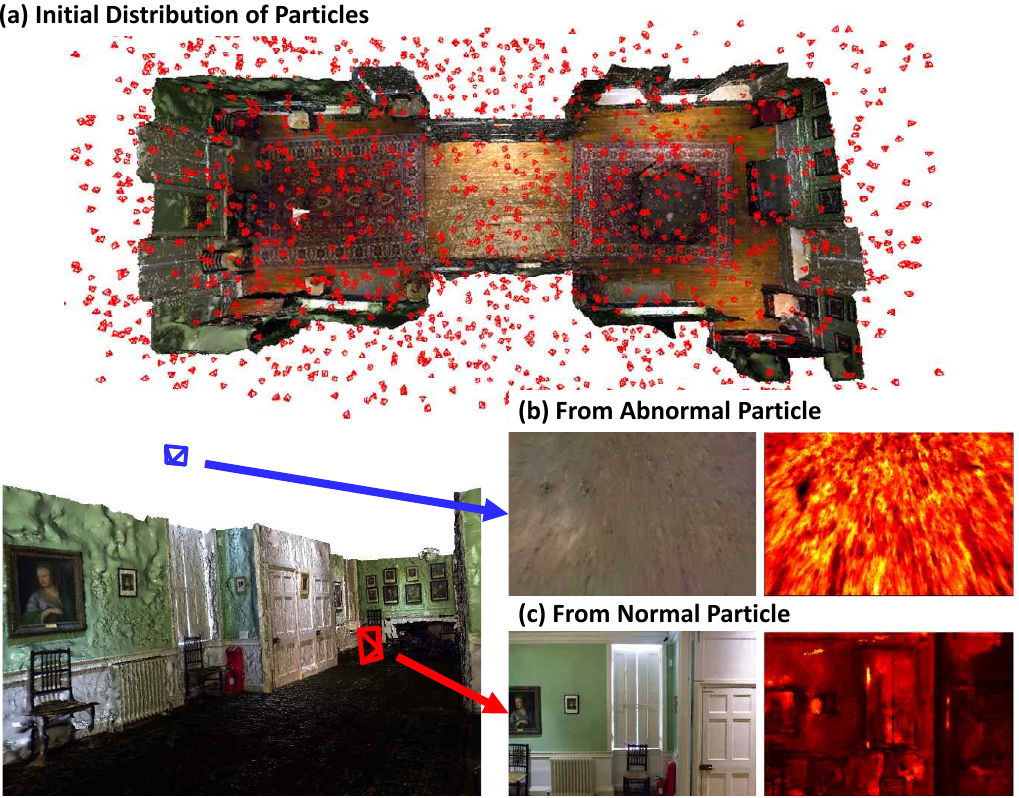}
\vspace{-5mm}
\caption{Our Fast Loc-NeRF utilzes particle uncertainty to estimates camera poses with observed image from NeRF map. (a) Initially distributed particles for global localization during the MCL process. (b) Rendered image and uncertainty map from an abnormal particle located outside the valid region. (c) Rendered image and uncertainty map from a normal particle.}
\label{fig:fig1}
\vspace{-5mm}
\end{figure}

Regarding the topic, several works have been reported. Gradient-based pose estimation methods, including iNeRF~\cite{yen2021inerf} and PI-NeRF~\cite{lin2023parallel}, optimize the camera pose by inverting the NeRF. APR-based approaches~\cite{chen2021direct,moreau2022lens,chen2022dfnet} utilize NeRF to train absolute pose regression networks to predict the pose directly. Among them, recently Loc-NeRF~\cite{maggio2023loc} was developed as a groundbreaking approach by integrating NeRF with Monte Carlo Localization (MCL), a probabilistic method traditionally used in mobile robotics for estimating the pose of a robot. 
The key idea of Loc-NeRF is to employ NeRF as a map model in the update step of the filter and robot dynamics for motion prediction. This integration allows for global localization and accurate tracking on NeRF, capitalizing on the high-quality environmental cues captured by NeRF. However, while Loc-NeRF demonstrated impressive localization capabilities, it also revealed a critical bottleneck: the computational intensity of rendering a vast number of rays across numerous particles within the MCL framework. This challenge becomes particularly critical in scenarios requiring real-time localization or operating under computational constraints.

To address this problem, we introduce a novel global localization method named Fast Loc-NeRF. A key idea in Fast Loc-NeRF is the particle rejection weighting scheme. This scheme estimates the uncertainty of each particle by leveraging the unique characteristics of NeRF’s rendering process and integrates this uncertainty into the particle weighting mechanism. It is actually the incorporation of rejection sampling into the importance sampling in MCL, improving the robustness of localization and accelerating the method, thereby contributing to the overall performance gains. Moreover, Fast Loc-NeRF proposes a coarse-to-fine approach, where matching between the rendered image and the observed image is performed at progressively higher resolutions as we move from coarse to fine phases. This approach allows the focus to shift from exploration to exploitation. As the resolution and the number of rendering pixels increase, the number of particles in MCL progressively decreases, keeping the computational load almost constant. The coarse phase is designed to explore as many candidates as possible with reasonable computational effort, while the fine phase is focused on intensively exploiting the most promising candidates to achieve the most accurate estimate.

Through extensive evaluations, we demonstrate that Fast Loc-NeRF sets new benchmarks in localization performance, offering both efficiency and precision across various standard datasets. Our contributions are aimed at advancing the application of NeRF in visual localization, providing a robust and efficient framework that paves the way for future research and practical implementations. 

\section{Related Work}
\label{sec: Related Works}

\textbf{Neural Radiance Fields (NeRF)} Neural Radiance Fields (NeRF)~\cite{mildenhall2021nerf} has emerged as a novel approach demonstrating that a neural network can be used to synthesize novel views of complex scenes. In detail, NeRF models the volumetric scene as a continuous function, using a deep network to map 5D coordinates (spatial location and viewing direction) to color and density. The success of NeRF has driven research aimed at more accurately representing environments, including addressing aliasing issues at various scales~\cite{barron2021mip,barron2022mip,barron2023zip}, achieving faster training and rendering times~\cite{muller2022instant,fridovich2022plenoxels,chen2022tensorf,sun2021direct}, and enabling the capability to learn larger scenes~\cite{turki2022mega,tancik2022block,mi2023switchnerf}, as well as other improvements~\cite{deng2022depth,bian2023nope,sandstrom2023point}. Additionally, it has extended into various applications such as SLAM~\cite{zhu2022nice,rosinol2023nerf,sucar2021imap}, editing~\cite{yuan2022nerf,mirzaei2023spin, Kong2023RoomNeRF}, dynamic scene reconstruction~\cite{pumarola2021d,fridovich2023k,li2023dynibar} and semantic understanding~\cite{zhi2021place,cen2023segment}.

\textbf{Visual Localization in NeRF} Visual localization, a crucial component in robotics and augmented reality, has been extensively studied, with methods ranging from particle-based approaches to optimization methods. The integration of NeRF into localization tasks represents a relatively new but rapidly growing area of research. The dense and continuous representation provided by NeRF offers a rich source of environmental cues that can be leveraged for more precise localization. Initial efforts in this space have focused on understanding how NeRF's environmental representations can be utilized to improve the accuracy of pose estimation. Several methods have emerged in this field, including gradient-based approaches~\cite{yen2021inerf,lin2023parallel}, APR-based methods~\cite{chen2021direct,moreau2022lens,chen2022dfnet}, and Loc-NeRF~\cite{maggio2023loc}, which utilizes Monte Carlo Localization. Gradient-based methods invert~\cite{yen2021inerf,lin2023parallel} NeRF to estimate pose by rendering images from NeRF and comparing them to actual images at the pixel level. They refine the camera's pose using gradients derived from this comparison. APR-based approaches~\cite{chen2021direct,moreau2022lens,chen2022dfnet} train absolute pose regression networks, such as PoseNet~\cite{kendall2015posenet}, to directly predict the pose, leveraging the renderable properties of NeRF. However, this method requires extensive training time and resources to train the APR network.
Loc-NeRF~\cite{maggio2023loc} is a notable example of combining NeRF with Monte Carlo Localization to achieve enhanced localization performance without any training process. However, the computational demands of this integration, particularly in the context of rendering multiple rays for numerous particles, present significant challenges. In this work, we proposed a novel approach named Fast-NeRF, which builds upon these foundations, addressing the computational challenges and enhancing the efficiency and accuracy of NeRF-based localization through coarse-to-fine localization framework and particle rejection weighting strategies.

\section{METHODOLOGY}
Our proposed method, Fast Loc-NeRF, is designed to address the computational challenges of NeRF-based localization, particularly focusing on the inefficiencies associated with the traditional Monte Carlo Localization approach when integrated with NeRF. Fast Loc-NeRF introduces a novel particle rejection weighting mechanism coupled with a coarse-to-fine localization framework, aimed at enhancing both the efficiency and accuracy of the localization process. 

\subsection{Monte Carlo Localization on NeRF}
\label{subsec:Monte Carlo Localization on NeRF}

\textbf{Neural Radiance Fields} Neural radiance fields(NeRF) learn the visual and geometry of the environment and represent a three-dimensional environment that enables novel view synthesis. Given a set of $n$ images $\mathcal{I} = \{ I_i \}_{i=1}^n$, and their corresponding camera poses and intrinsic parameters, NeRF encodes a function $F_{\textit{NeRF}} : (x,d) \rightarrow (c,\sigma)$ from 3D coordinate $x = (x,y,z)$ and viewing direction $d = (\theta,\phi)$ to RGB color $c$ and density $\sigma$. The color and the densities of sampled points along the ray are predicted and integrated to render a novel view. 

\begin{equation}
    \hat{\mathrm{C}}(r) = \int_0^\infty T(z) \sigma(z)\mathrm{c}(z)dt,
\end{equation}
where $T(z) = \mathrm{exp}(-\int_0^z\sigma(s)ds)$ is the transmittance checking occlusion by integrating along the ray $r(z) = o+ zd$ between 0 to $z$. Since the density and radiance values are derived from the 5D coordinate of points on rays, NeRF techniques employ a sampling-based Riemann summation to approximate the integral. NeRF optimizes by the reconstruction error between the rendered colors and the corresponding ground-truth colors. 
\begin{equation}
   L_{color} = \sum_{r \in \mathcal{R}} \lVert\hat{\mathrm{C}}(r) - \mathrm{C}(r) \rVert^2,
\end{equation}
where $\mathcal{R}$ is the set of rays $r$ with the known camera poses. 

\begin{figure*}[htbp]
\centering
\includegraphics[width=0.95\textwidth]{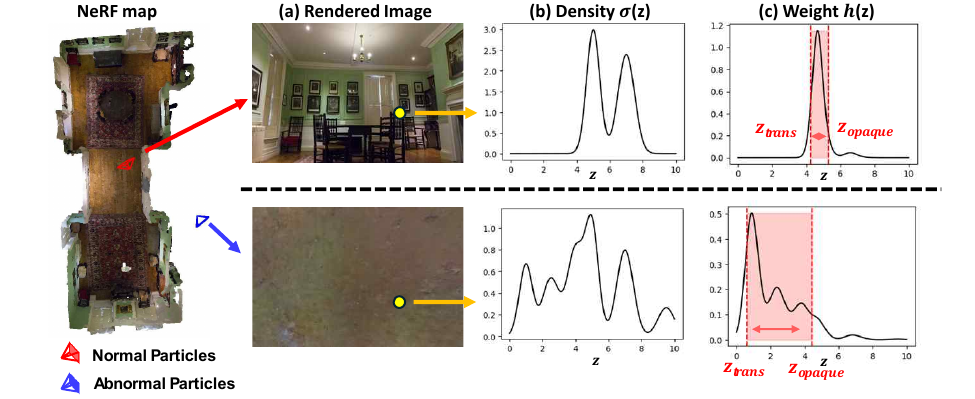}
\vspace{-3mm}
\caption{\textbf{Particle Rejection Weighting} We illustrate the particle rejection weighting process. (a) Rendered image from the red and blue particle each, which is located inside and outside room (b) Density distribution $\sigma(z)$ over ray  $r(z)=o+zd$ (c) Ray weight distribution $h(z)=T(z)\sigma(z)$ and red region indicate the range between $z_{\text{trans}}$ and $z_{\text{opaque}}$}
\vspace{-5mm}
\label{fig:fig3} 
\end{figure*}

 \textbf{Monte Carlo Localization leveraging NeRF as a map} With a well-trained Neural Radiance Field (NeRF) map denoted as \(\mathcal{M}\), RGB images \(I_t\) representing observations at time \(t\), and motions \(O_t\) at time \(t\), a particle filter utilizing Monte Carlo localization is employed to estimate the 6 Degrees of Freedom (DoF) camera pose \(X_t\) for the corresponding time \(t\). This process uses the particle filter to calculate the posterior probability \(P(X_t|\mathcal{M},I_{1:t},O_{1:t})\), incorporating all images \(I_{1:t}\) and motions \(O_{1:t}\) recorded from the initial moment up to time \(t\). In the particle filter, a weighted set of \(n\) particles represent the distribution of possible camera poses as follows:

\begin{equation}
    \mathcal{S}_t = \{ \langle x_t^{[i]}, w_t^{[i]} \rangle \mid i = 1, \ldots, N \},
\end{equation}

where \(x_t^{[i]}\) is the 6DoF pose of the \(i\)th particle and \(w_t^{[i]}\) is the associated weight of \(x_t^{[i]}\) normalized from 0 to 1. The particle filter updates the group of particles every time new images and measurements are observed.
During the prediction step, the set of particles \(\mathcal{S}_t\) at time \(t\) is estimated from the set of particles \(\mathcal{S}_{t-1}\) with the motion of the camera \(O_t\) between time \(t\) and time \(t-1\) as the motion model \(P(X_t|X_{t-1}, O_t)\). The pose \(x_t^{[i]}\) of each particle is sampled through Gaussian noise as follows $x_t^{[i]} =  x_{t-1}^{[i]} \cdot O_t \cdot x_\epsilon^{[i]},$ where \(x_\epsilon \in \mathbb{SE}(3)\) is the prediction noise modeled as an exponential map \(\text{Exp}(\delta)\).
Then, the importance weights of particles are updated based on the measurement likelihood \(P(I_t|X_t, \mathcal{M})\), which indicates the probability of the pose \(x_t^{[i]}\) given the observed image \(I_t\) at time \(t\). Loc-NeRF~\cite{maggio2023loc} employed a heuristic function to estimate the measurement likelihood as follows:
\begin{equation}
w_t^{[i]} = \left( \frac{B}{\sum_{j=1}^{B}  \left( I_t(p) - \hat{C}(r(p, x_t^{[i]}), K) \right)^2} \right)^4,   
\end{equation}
where \(B\) denotes the number of pixels rendered per particle. \(\hat{C}(r(p_j, x_t^{[i]}), K)\) is the rendered color on the pixel \(p_j\) with the pose \(x_t^{[i]}\) and shared intrinsic camera parameter \(K\). After importance weighting, \(N\) particles are resampled with normalized weights \(w_t^{[i]}\).

\subsection{Particle Rejection Weighting}
\label{sec:sec3_2}
\textbf{Motivation} When we apply a vanilla MCL to a NeRF map, we often encounter situations where early-stage particles are randomly or uniformly distributed throughout the environment, and some of them fall into regions where the NeRF is not trained at all. This occurs because we have no information about the structure of the environment. An example of this situation is shown in the leftmost figure in Fig \ref{fig:fig3}. In the figure, the red particle (in the first row) falls into the “valid” region where the NeRF map is trained and successfully renders the corresponding image from the NeRF map. In contrast, the blue particle (in the second row) happens to fall outside the environment (“invalid region”) where the NeRF map is not trained at all. As a result, the blue particle renders a strange image, and such abnormal particles should be removed from subsequent processing. In this paper, we remove such abnormal particles by incorporating rejection sampling into the importance sampling framework.
While importance sampling and rejection sampling are distinct techniques, they are related in that both methods use the ratio of the target distribution to the proposal distribution to draw samples. Specifically, we assign very low importance weights to particles that should be removed from further consideration.

\textbf{Particle Weighting} The remaining problem is how to compute the importance weight so that the abnormal particles are removed in the context of importance sampling. When a normal particle located in the valid region renders an image by querying a NeRF with rays, the density $\sigma(z)$ is low for empty spaces but abruptly increases for objects such as walls, as shown in the first row of Fig \ref{fig:fig3}~(b). In this case, if we define $z_{\text{trans}}$ and $z_{\text{opaque}}$  as the points on the ray until the ray remains transparent and the point from which the ray becomes opaque, respectively, and compute them by: 
\begin{equation}
\begin{split}
    z_{\text{trans}} &= \min \left\{ z \mid \int_{0}^{z} T(s) \sigma(s) \, ds \geq \alpha \right\},\\
    z_{\text{opaque}} &= \max \left\{ z \mid \int_{0}^{z} T(s) \sigma(s) \, ds \leq 1-\alpha \right\}.
\end{split}
\label{eq:equation5}
\end{equation}

The difference between the two points $z_{\text{opaque}}$ and $z_{\text{trans}}$ is relatively small, as shown in the first row of Fig \ref{fig:fig3} (c), where $\alpha$ is a predefined threshold close to zero. On the contrary, when the abnormal particle located on the invalid region renders an image by querying a NeRF with rays, the density $\sigma(z)$ tends to have meaningless values, lacking a clear distinction between the empty space and the object, as shown in the second row of Fig \ref{fig:fig3} (b). Consequently, the corresponding ray weight distribution $h(z)=T(z)\sigma(z)$ does not have a clear distinction between the transparent point $z_{\text{trans}}$ and the opaque point $z_{\text{opaque}}$, as shown in the second row of Fig \ref{fig:fig3} (c). In this case, the difference between the two points $z_{\text{opaque}}$ and $z_{\text{trans}}$ is relatively large. To remove such abnormal particles in the context of importance sampling, we weight each particle by:

\begin{equation}
\label{eq:equation6}
\begin{split}
    E^{R_t}_{color} &= \left( I^{R_t}_t(p_j) - \hat{C}(r(p_j, x^{[i]}_t), R_t \cdot K) \right)^,\\
    w^{[i]}_t &= \left( \frac{B_t}{\sum_{j=1}^{B_t} E^{R_t}_c \cdot F(r(p_j, x^{[i]}_t))} \right)^4,
\end{split}
\end{equation}


where $F(r(p_j)) = \max(z_{\text{opaque}} - z_{\text{trans}}, \tau)$ penalizes the abnormal particles using the gap between the transparent point $z_{\text{trans}}$ and the opaque point $z_{\text{opaque}}$. $\tau$ is a minimum bound of $F(r(p_j))$, and it simply ensures the algorithm stability.

\subsection{Coarse-to-Fine Multi-Scale Matching}
\label{sec:sec3_3}

\begin{figure}[htbp]
\centering
\includegraphics[width=0.5\textwidth]{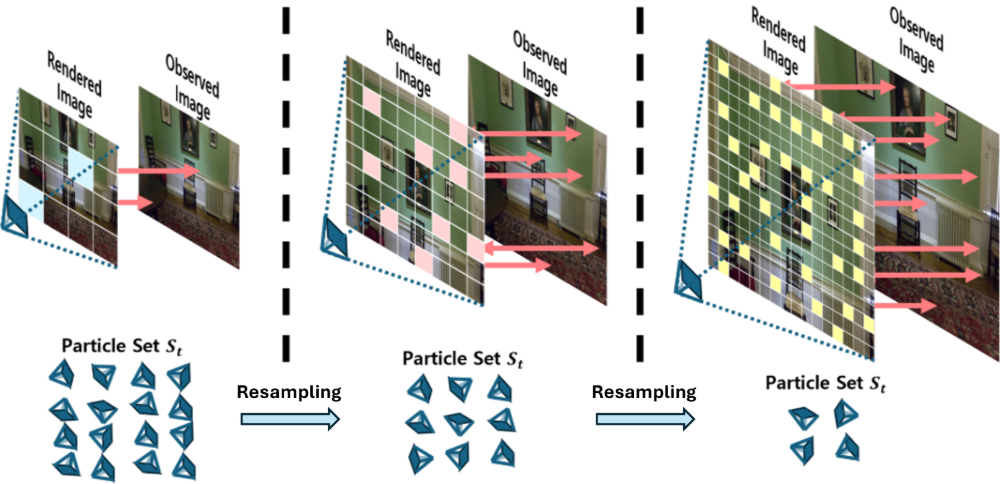}
\caption{Overview of our Coarse-to-Fine Multi-Scale Matching. Our approach shifts focus from exploration to exploitation, increasing the resolution and the number of rendering pixels while we progressively decreases the number of particles in MCL.}
\label{fig:fig1}
\end{figure}

\textbf{Framework} Our Fast Loc-NeRF dynamically adjusts the localization process through a gradual refinement of computation, transitioning from a coarse exploration phase to a fine exploitation phase. The key idea of Fast Loc-NeRF is to maintain the amount of computation represented by the product of the number of particles $N$ and the number of probings $B$ (renderings or queries) almost constant in all phases, achieving both efficiency and precision. Specifically, as the filtering steps progress from coarse to fine, the focus shifts from utilizing a large number of particles $N$ to increasing the number of query pixels $B$ for precise localization. This adjustment narrows down the search space and focuses on the most promising solutions. During the coarse phase, fewer renderings challenge the capture of the image's overall shape, so we adjust the rendering resolution $R$ in each phase to maintain the matching region's proportion over the image. In the algorithm~\ref{alg:mcl-nerf}, $R$, $B$, and $N$ denote the rendering scale, the number of queries, and the number of particles, respectively. As we move to the fine phase, we increase the rendering scale $R$ from $\sfrac{1}{16}$ resolution to full resolution, while also increasing the number of query rays $B$ in accordance with the resolution.

\begin{algorithm}
 \SetCustomAlgoRuledWidth{0.5\textwidth}
  \caption{MCL-NeRF}
  \label{alg:mcl-nerf}
  \KwIn{$X_{t-1}, u_t, I_t, R_t, B_t,N_t$}
  $\Bar{X}_t = X_t = \emptyset$\\
  \For{$i = 1$ to $N_{t-1}$}{
    sample $x_t^{[i]} \sim p(x_t \mid u_t, x_{t-1}^{[i]})$\\
    $I_{t}^{R_t} \leftarrow \text{downscale}(I_t, R_t)$\\
    \For{$j = 1$ to $B_t$}{
        render $\hat{C}(r(p_j, x_t^{[i]}), R_t)$
    }
    $w_t^{[i]} = p(I_{t}^{R_t} \mid x_t^{[i]},\mathcal{M})$\\
    calculate $w_t^{[i]}$ through Eq \eqref{eq:equation6} \\
    $\Bar{X}_t = \Bar{X}_t + \{\langle x_t^{[i]}, w_t^{[i]} \rangle\}$\\
  }
  \For{$i = 1$ to $N_t$}{
    draw $i$ with probability $\propto w_t^{[i]}$\\
    add $x_t^{[i]}$ to $X_t$
  }
  \textbf{Adjust} $R_{t+1}, B_{t+1},N_{t+1}$ \\
  \Return $X_t,R_{t+1}, B_{t+1},N_{t+1}$
\end{algorithm}
\vspace{-5mm}

\textbf{Detailed Structure} 
Fast Loc-NeRF operates by gradually refining the particle distribution through two stages of refinement, shifing from coarse exploration to precise localization. The process begins with an initial particle distribution at $t=0$ and undergoes 2 stages of refinement, where each stage is triggered when the particles converge and the variance of their 3D positions \( \{x_t^{[i]}\} \), fall within specific thresholds, \(\delta_{\text{refine}}^{1,2}\), as \( \text{Var}(\{x_t^{[i]}\}) < \delta_{\text{refine}}^{1,2} \). With each threshold, the algorithm refines the search space further by decreasing the number of particles and increasing the rendering resolution, ensuring progressively more accurate localization. In each refinement step, MCL-NeRF (as described in Algorithm \ref{alg:mcl-nerf}) is used to update the particles through sampling, importance weighting, and resampling. In MCL-NeRF, we downscale the observed image \(I_t\) to \(I_t^{R_t}\), and render the image \(\hat{C}(r(p_j, x_t^{[i]}), R_t \cdot K)\) of query points \({\{p_j\}}_{j=1}^{B_t}\) , which is used to compute the importance weights $w^{[i]}$. These two stages of refinement ensure that Fast Loc-NeRF balances efficiency and precision.

\begin{table*}[tb]
\centering
\small
\begin{tabularx}{0.93\textwidth}{c|cccccccc|c}
\toprule 

Method & Fern & Flower & Leaves & Horns & Fortress & Room & Orchids & Trex & Time \\ \midrule 

iNeRF~\cite{yen2021inerf}   & 0.75/167 & 0.92/75.0 & 1.07/145 & 1.12/138 & 1.26/136 & 1.47/111 & 1.22/152 & 1.06/132 & \textbf{0.208}\\
PI-NeRF~\cite{lin2023parallel}   & 1.02/103 & 0.98/96.8 & 0.97/101 & 1.07/93.7 & 1.03/92.4 & 1.05/96.9 & 1.00/98.6 & 1.08/70.0 & - \\
Loc-NeRF~\cite{maggio2023loc}   & 0.03/0.64 & 0.56/18.1 & 0.89/4.58 & 0.39/3.84 & 0.03/\bf{0.96} & 0.06/0.94 & 1.43/35.0 & 0.64/5.68 & 0.724 \\
Ours    & \bf{0.02}/\bf{0.59} & \bf{0.01}/\bf{0.19} & \bf{0.21}/\bf{0.72} & \bf{0.02}/\bf{0.23} & \textbf{0.03}/1.01 & \bf{0.05}/\bf{0.94} & \bf{0.38}/\bf{3.47} & \bf{0.14}/\bf{1.81} & 0.327 \\ \bottomrule 
\end{tabularx}
\caption{Comparison Result of Global Localization on the LLFF dataset, with position error (cm), rotation error (deg), and time for each update step (sec).}
\vspace{-8mm}
\label{tab:table1}
\end{table*}

\begin{figure*}[htbp]
\centering
\includegraphics[width=\textwidth]{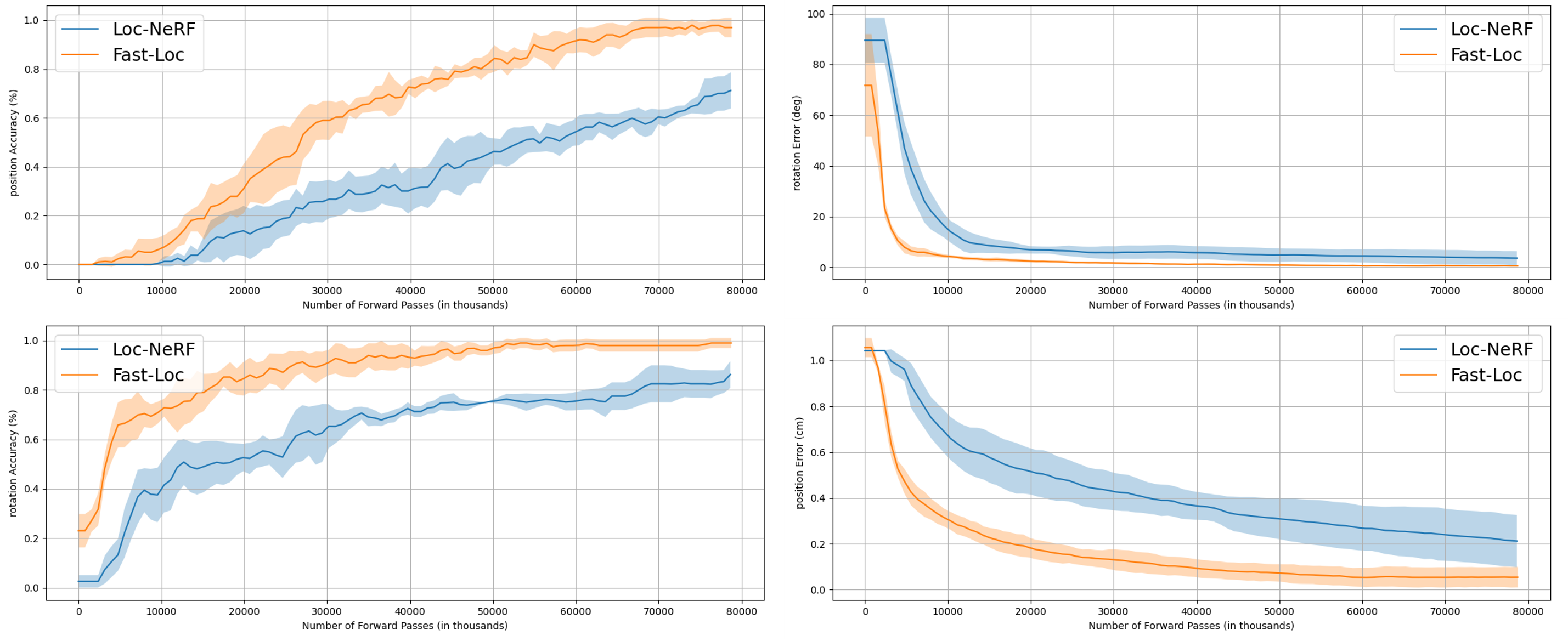}
\vspace{-7mm}
\caption{Quantitative comparison with Loc-NeRF. The first row presents position accuracy (\%) and position error (cm). The second row shows rotation accuracy (\%) and error (cm) on four datasets from LLFF, which are the same datasets conducted on Loc-NeRF paper.}
\vspace{-5mm}
\label{fig:fig5}
\end{figure*}

\section{Experiments}
\label{sec:sec4}
Section~\ref{subset:4.1} focuses on benchmarking Fast Loc-NeRF against Loc-NeRF~\cite{maggio2023loc}, iNeRF \cite{yen2021inerf}, and PI-NeRF (Parallel Inversion of NeRF)~\cite{lin2023parallel} in global localization settings to demonstrate its superior accuracy and speed. Section~\ref{subset:4.2} explores the benefits of incorporating multi-scale representation, utilizing Zip-NeRF~\cite{barron2023zip} as a map. We expect the synergy between multi-scale matching and multi-scale representation to enable more accurate particle weighting. Ablation studies for module analysis, particle rejection weighting and the number of initial particles are demonstrated on Section~\ref{subset:4.3}.

\subsection{Global Localization}
\label{subset:4.1}
\textbf{Experiment Setting} For our evaluation, we conducted global localization experiments across all 8 forward-facing scenes of the LLFF dataset~\cite{mildenhall2019local} on a workspace equipped with RTX 2080Ti and 3GHz Intel Core i5-12500. We generated maps using a basic NeRF model~\cite{mildenhall2021nerf} with NeRF-pytorch~\cite{lin2020nerfpytorch} same as comparisons, and trained for 20k iterations to represent each scene accurately. We followed the initial distribution settings of Loc-NeRF~\cite{maggio2023loc}, generating offset pose from the ground truth pose within a range of [-1m, 1m] for each axis. In comparison to our method, Loc-NeRF was configured with a batch size (ray per particle) of 32 and utilized 600 particles. We also applied $particle$ $annealing$, reducing the number of particles to 100. In the case of iNeRF~\cite{yen2021inerf} and PI-NeRF~\cite{lin2023parallel}, the experimental setting follows the original methodology of each paper. The offset pose is used as the initial pose for the experiments on gradient-based optimization comparisons. For our setting, the threshold $\tau$ for particle rejection weighting is 0.1. Observed images were downscaled to $R=\{\sfrac{1}{4},\sfrac{1}{2},1\}$, the number of particles is $N=\{9600,600,100\}$ and the number of rendered rays are $B=\{8,16,32\}$ for coarse-to-fine multi-scale matching.

\textbf{Result} This section compares Fast Loc-NeRF, Loc-NeRF~\cite{maggio2023loc}, and gradient-based methods~\cite{yen2021inerf,lin2023parallel} in terms of position and rotation accuracy over the number of forward passes, as shown in Table~\ref{tab:table1}. Gradient-based methods generally struggle with convergence from initial poses. In contrast, Loc-NeRF~\cite{maggio2023loc}, employing Monte Carlo Localization, often accurately estimates the true position and rotation. Fast Loc-NeRF enhances this efficiency, achieving comparable accuracy more quickly and showing significant improvements in average time per update step and final accuracy. Fig~\ref{fig:fig5} shows the qualitative results for the four scenes ($\textit{Fern}$, $\textit{Fortress}$, $\textit{Horns}$, and $\textit{Room}$) of LLFF which are same scenes experimented on Loc-NeRF~\cite{maggio2023loc}. Pose accuracy is defined as a position error within 5 cm and a rotation error under 5 degrees.

\subsection{Localization on Multi-Scale Representation}
\label{subset:4.2}

\begin{table*}[ht]{
\caption{Comparison on LLFF dataset~\cite{mildenhall2019local} and Deep Blender dataset~\cite{hedman2018deep} utilizing Zip-NeRF~\cite{barron2023zip} as a map.}
\vspace{-3mm}
\label{tab:tab2}
\centering

\resizebox{1.0\linewidth}{!}{
\begin{tabular}{@{}cccccccccccc@{}}
\toprule
 & \multicolumn{8}{c}{(a) LLFF} & \multicolumn{2}{c}{(b) Deep Blending} & \\
 & Fern       & Leaves    & Flower     & Horns      & Fortress  & Room       & Orchids     & Trex  & Drjohnson & Playroom    &  Time   \\ 
 \cmidrule(r){2-9} \cmidrule(r){10-11} \cmidrule(r){12-12} 
  \multicolumn{12}{c}{Position Error (cm) / Rotation Error ($\circ$)}\\
  \addlinespace
Loc-NeRF & 1.54/57.00 & 0.63/14.5 & 0.37/33.00 & 0.64/12.00 & 0.28/1.59 & 0.34/5.60  & \textbf{1.10}/133.00 & 0.54/4.45 & 0.83/54.40 & 0.97/35.65 & 0.205 \\
Ours & \bf{0.67}/\bf{16.40} & \bf{0.32}/4.30 & \bf{0.10}/\bf{21.00} & \bf{0.02}/\bf{0.22}  & \bf{0.02}/\bf{0.49} & \textbf{0.02} /\textbf{0.38}  & 1.80/\bf{84.00}  & \bf{0.06}/\bf{0.73} & \bf{0.42}/\bf{34.04} & \bf{0.37}/\bf{33.40} &\textbf{0.118} \\  \cmidrule(r){2-9} \cmidrule(r){10-11} \cmidrule(r){12-12} 
 \multicolumn{12}{c}{Position Error $<$ 5cm / Rotation Error $<5^{\circ}$ }\\
 \addlinespace
Loc-NeRF &  0.00/0.00 & 0.10/0.70 & 0.55/0.65 & 0.50/0.75 & 0.55/0.95 & 0.35/0.60 & 0.00/0.00 & 0.70/0.75 & 0.10/0.10 & 0.10/0.15 & 0.205 \\
Ours & \textbf{0.68}/\textbf{0.68} & \textbf{0.28}/\textbf{0.80} & \textbf{0.88}/\textbf{0.88} & \textbf{1.00}/\textbf{1.00} & \textbf{1.00}/\textbf{1.00} & \textbf{0.96}/\textbf{1.00} & \textbf{0.20}/\textbf{0.20} & \textbf{0.80}/\textbf{0.96} & \bf{0.35}/\bf{0.25} & \bf{0.25}/\bf{0.40} & \textbf{0.118} \\ \bottomrule
\end{tabular}
}
\vspace{-3mm}
}
\end{table*}

\textbf{Experiment Setup}: We utilized multi-scale representation via Zip-NeRF~\cite{barron2023zip} to conduct comparative analyses between Loc-NeRF and Fast Loc-NeRF. The experimental dataset consists of all 8 scenes from the LLFF dataset~\cite{mildenhall2019local}, consistent with Section~\ref{subset:4.1}, and two additional scenes from the Deep Blending dataset~\cite{hedman2018deep}. 
The Deep Blending dataset~\cite{hedman2018deep}, unlike the LLFF dataset~\cite{mildenhall2019local} which comprises primarily forward-facing scenes, showcases entire rooms, making it particularly suitable for demonstrating global localization for the full real world.

\textbf{Results} We present the results of employing Zip-NeRF~\cite{barron2023zip} as a map on our method and Loc-NeRF in Table~\ref{tab:tab2}. Our approach, Fast Loc-NeRF demonstrated superior performance against baseline across most of the forward-facing scenes. While Loc-NeRF~\cite{maggio2023loc} had difficulty reaching a accuracy at 80\% in any scene, our Fast Loc-NeRF method consistently attained this benchmark across most of the scenes. On the Deep Blending dataset, each scene representing a single room is a more challenging environment for global localization compared to the forward-facing scenes. Our method significantly outperforms Loc-NeRF shown in Table~\ref{tab:tab2}. This enhancement suggests that our method efficiently explores extensive spatial environments.

\subsection{Ablation Studies}
\label{subset:4.3}

\textbf{Module Ablation} Fast Loc-NeRF proposes two methods: a particle rejection weighting and a coarse-to-fine localization framework. Table~\ref{tab:tab3} shows the results of applying each method separately. 
By applying a coarse-to-fine localization framework, the overall time per update step is halved while maintaining localization accuracy as shown in Table~\ref{tab:tab3}. Additionally, the implementation of particle rejection weighting not only improves position accuracy but also accelerates convergence by assigning lower weights to abnormal particles, leading to a reduction in overall computation time.

\begin{table}[ht]
\centering
\small
\resizebox{1.0\linewidth}{!}{
\begin{tabular}{@{}c|cc|cc|c@{}}
\toprule
  & \multicolumn{2}{c|}{Position} & \multicolumn{2}{c|}{Rotation} & \multirow{2}{*}{Time per  step} \\
  & Accuracy & Error & Accuracy & Error &   \\
\midrule
w/o Both     & 52.5 & 0.50 & 65.0 & 5.84 & 0.7450 \\
w/o Rejection  & 55.0 & 0.45 & 75.0 & 3.18 & 0.3836 \\
w/o Coarse-to-Fine  & 65.0 & 0.33 & 85.0 & 8.12 & 0.5056 \\
Full  & \textbf{75.0} & \textbf{0.11} & \textbf{97.5} & \textbf{1.09} & \textbf{0.3267} \\ \bottomrule
\end{tabular}
}
\caption{Results of the module ablation on Fast Loc-NeRF.}
\vspace{-5mm}
\label{tab:tab3}
\end{table}

\begin{figure}[tb]
    \centering
    \includegraphics[width=0.9\columnwidth]{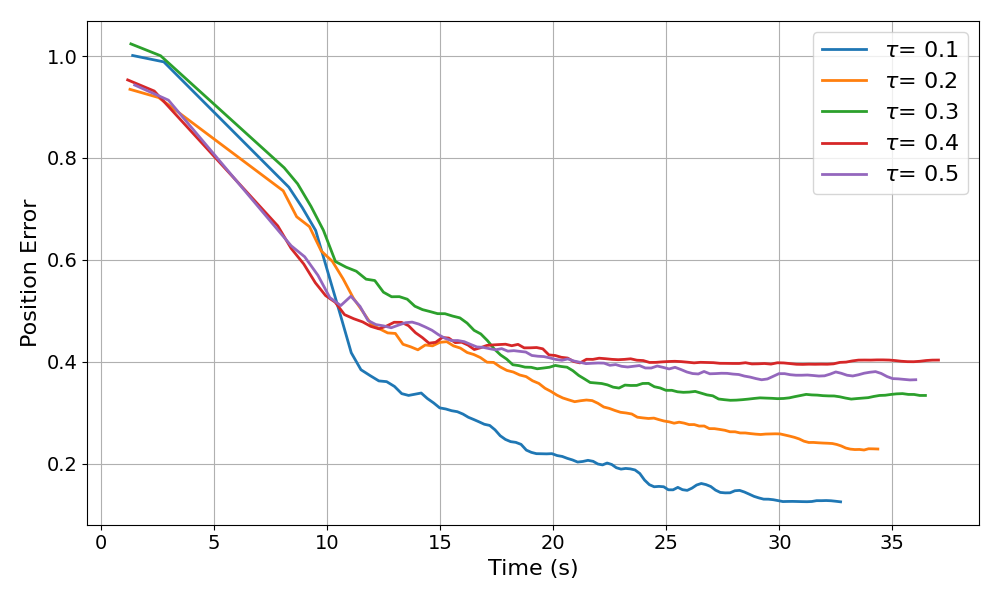}
    \vspace{-5mm}
    \caption{Ablation studies on threshold of particle rejection weighting.}
    \label{fig:ablation_rejection}
    \vspace{-5mm}
\end{figure}

\textbf{Ablation Study on Particle Rejection Weighting} Our particle rejection weighting method stabilizes by setting a minimum bound $\tau$ on the gap between the last transparent point $z_{\text{trans}}$ and the opaque point $z_{\text{opaque}}$ as $F(r(p_j)) = \max(z_{\text{opaque}} - z_{\text{trans}}, \tau)$ on Equation~\ref{eq:equation6}. Fig~\ref{fig:ablation_rejection} presents the average of localization results on 8 scenes of LLFF dataset for various minimum bounds $\tau$. As observed from Fig~\ref{fig:ablation_rejection}, smaller values of $\tau$ make the particle rejection weighing more effective in reducing position errors. Additionally, when $\tau$ exceeds 0.4, the results become similar to those obtained without particle rejection weighting. Since the minimum value becomes $\tau$ large, the uncertainty no longer significantly influences the importance weighting process. 

\begin{figure}[!h]
    \centering
    
    \includegraphics[width=0.9\columnwidth]{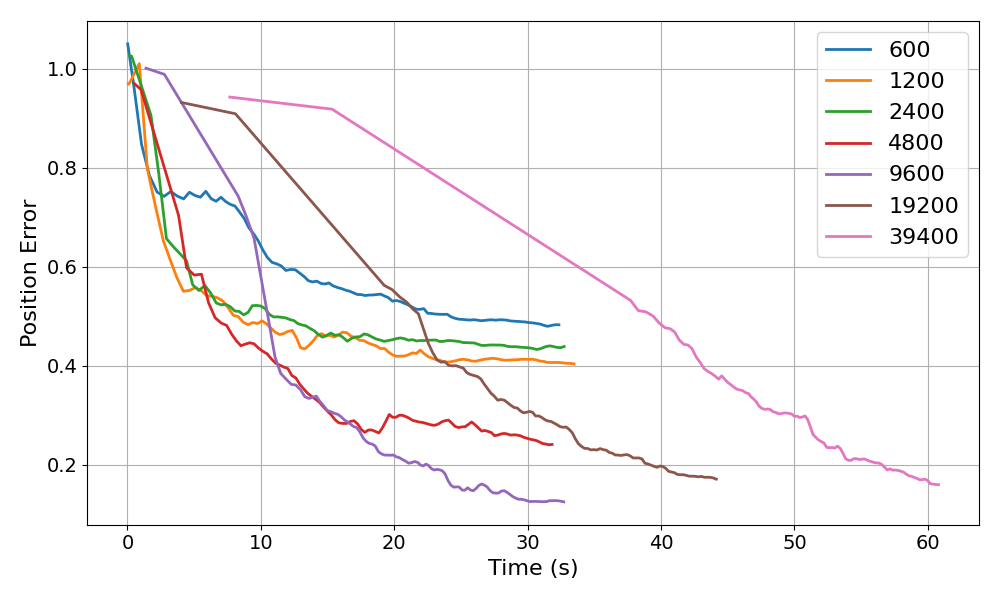}
    \vspace{-5mm}
    \caption{Results with various numbers of initial particles on our framework.}
    \label{fig:ablation_initial}
    \vspace{-2mm}
\end{figure}

\textbf{Ablation Study on \# of initial particle} Since no prior information is provided on the NeRF map $\mathcal{M}$, we initially sample the particles randomly or uniformly across the searching region. Fig~\ref{fig:ablation_initial} illustrates the results of the global localization with varying numbers of initial particles. Increasing the number of initial particles improves exploration, resulting in lower position error and higher accuracy.  However, beyond a certain threshold, the inference time for localization during the coarse phase increases exponentially.

\section{Conclusion}

In summary, we introduce Fast Loc-NeRF, which enhances efficiency and accuracy in NeRF map-based global localization. To deal with particles distributed in irrelevant locations, we propose particle rejection weighting, which estimates ray uncertainty by analyzing the distribution over each ray without any additional training step. By not simply using the RGB values obtained from rendering, but also applying the uncertainty of each ray, we reduced the exploration of inappropriate space, enabling faster and more accurate localization. Furthermore, we exploit particle filtering in a coarse-to-fine manner to enable efficient localization on a NeRF map. While progressively matching particle states and measurements from low to high resolution, Fast Loc-NeRF excelerates the expansive particle weight update process. As a result, our method speeds up the localization process and improves localization accuracy by evaluating various NeRF datasets and multi-scale NeRF models.

\section*{Acknowledgements}
This work was supported by Korea Evaluation Institute Of Industrial Technology (KEIT) grant funded by the Korea government(MOTIE) (No.20023455, Development of Cooperate Mapping, Environment Recognition and Autonomous Driving Technology for Multi Mobile Robots Operating in Large-scale Indoor Workspace)








\bibliographystyle{IEEEtran}
\bibliography{ref}

\begin{thebibliography}{10}
\providecommand{\url}[1]{#1}
\csname url@samestyle\endcsname
\providecommand{\newblock}{\relax}
\providecommand{\bibinfo}[2]{#2}
\providecommand{\BIBentrySTDinterwordspacing}{\spaceskip=0pt\relax}
\providecommand{\BIBentryALTinterwordstretchfactor}{4}
\providecommand{\BIBentryALTinterwordspacing}{\spaceskip=\fontdimen2\font plus
\BIBentryALTinterwordstretchfactor\fontdimen3\font minus \fontdimen4\font\relax}
\providecommand{\BIBforeignlanguage}[2]{{%
\expandafter\ifx\csname l@#1\endcsname\relax
\typeout{** WARNING: IEEEtran.bst: No hyphenation pattern has been}%
\typeout{** loaded for the language `#1'. Using the pattern for}%
\typeout{** the default language instead.}%
\else
\language=\csname l@#1\endcsname
\fi
#2}}
\providecommand{\BIBdecl}{\relax}
\BIBdecl

\bibitem{mur2017orb}
R.~Mur-Artal and J.~D. Tard{\'o}s, ``Orb-slam2: An open-source slam system for monocular, stereo, and rgb-d cameras,'' \emph{IEEE transactions on robotics}, vol.~33, no.~5, pp. 1255--1262, 2017.

\bibitem{campos2021orb}
C.~Campos, R.~Elvira, J.~J.~G. Rodr{\'\i}guez, J.~M. Montiel, and J.~D. Tard{\'o}s, ``Orb-slam3: An accurate open-source library for visual, visual--inertial, and multimap slam,'' \emph{IEEE Transactions on Robotics}, vol.~37, no.~6, pp. 1874--1890, 2021.

\bibitem{grisetti2007improved}
G.~Grisetti, C.~Stachniss, and W.~Burgard, ``Improved techniques for grid mapping with rao-blackwellized particle filters,'' \emph{IEEE transactions on Robotics}, vol.~23, no.~1, pp. 34--46, 2007.

\bibitem{mildenhall2021nerf}
B.~Mildenhall, P.~P. Srinivasan, M.~Tancik, J.~T. Barron, R.~Ramamoorthi, and R.~Ng, ``Nerf: Representing scenes as neural radiance fields for view synthesis,'' \emph{Communications of the ACM}, vol.~65, no.~1, pp. 99--106, 2021.

\bibitem{yen2021inerf}
L.~Yen-Chen, P.~Florence, J.~T. Barron, A.~Rodriguez, P.~Isola, and T.-Y. Lin, ``inerf: Inverting neural radiance fields for pose estimation,'' in \emph{2021 IEEE/RSJ International Conference on Intelligent Robots and Systems (IROS)}.\hskip 1em plus 0.5em minus 0.4em\relax IEEE, 2021, pp. 1323--1330.

\bibitem{lin2023parallel}
Y.~Lin, T.~M{\"u}ller, J.~Tremblay, B.~Wen, S.~Tyree, A.~Evans, P.~A. Vela, and S.~Birchfield, ``Parallel inversion of neural radiance fields for robust pose estimation,'' in \emph{2023 IEEE International Conference on Robotics and Automation (ICRA)}.\hskip 1em plus 0.5em minus 0.4em\relax IEEE, 2023, pp. 9377--9384.

\bibitem{chen2021direct}
S.~Chen, Z.~Wang, and V.~Prisacariu, ``Direct-posenet: Absolute pose regression with photometric consistency,'' in \emph{2021 International Conference on 3D Vision (3DV)}.\hskip 1em plus 0.5em minus 0.4em\relax IEEE, 2021, pp. 1175--1185.

\bibitem{moreau2022lens}
A.~Moreau, N.~Piasco, D.~Tsishkou, B.~Stanciulescu, and A.~de~La~Fortelle, ``Lens: Localization enhanced by nerf synthesis,'' in \emph{Conference on Robot Learning}.\hskip 1em plus 0.5em minus 0.4em\relax PMLR, 2022, pp. 1347--1356.

\bibitem{chen2022dfnet}
S.~Chen, X.~Li, Z.~Wang, and V.~A. Prisacariu, ``Dfnet: Enhance absolute pose regression with direct feature matching,'' in \emph{European Conference on Computer Vision}.\hskip 1em plus 0.5em minus 0.4em\relax Springer, 2022, pp. 1--17.

\bibitem{maggio2023loc}
D.~Maggio, M.~Abate, J.~Shi, C.~Mario, and L.~Carlone, ``Loc-nerf: Monte carlo localization using neural radiance fields,'' in \emph{2023 IEEE International Conference on Robotics and Automation (ICRA)}.\hskip 1em plus 0.5em minus 0.4em\relax IEEE, 2023, pp. 4018--4025.

\bibitem{barron2021mip}
J.~T. Barron, B.~Mildenhall, M.~Tancik, P.~Hedman, R.~Martin-Brualla, and P.~P. Srinivasan, ``Mip-nerf: A multiscale representation for anti-aliasing neural radiance fields,'' in \emph{Proceedings of the IEEE/CVF International Conference on Computer Vision}, 2021, pp. 5855--5864.

\bibitem{barron2022mip}
J.~T. Barron, B.~Mildenhall, D.~Verbin, P.~P. Srinivasan, and P.~Hedman, ``Mip-nerf 360: Unbounded anti-aliased neural radiance fields,'' in \emph{Proceedings of the IEEE/CVF Conference on Computer Vision and Pattern Recognition}, 2022, pp. 5470--5479.

\bibitem{barron2023zip}
------, ``Zip-nerf: Anti-aliased grid-based neural radiance fields,'' \emph{arXiv preprint arXiv:2304.06706}, 2023.

\bibitem{muller2022instant}
T.~M{\"u}ller, A.~Evans, C.~Schied, and A.~Keller, ``Instant neural graphics primitives with a multiresolution hash encoding,'' \emph{ACM Transactions on Graphics (ToG)}, vol.~41, no.~4, pp. 1--15, 2022.

\bibitem{fridovich2022plenoxels}
S.~Fridovich-Keil, A.~Yu, M.~Tancik, Q.~Chen, B.~Recht, and A.~Kanazawa, ``Plenoxels: Radiance fields without neural networks,'' in \emph{Proceedings of the IEEE/CVF Conference on Computer Vision and Pattern Recognition}, 2022, pp. 5501--5510.

\bibitem{chen2022tensorf}
A.~Chen, Z.~Xu, A.~Geiger, J.~Yu, and H.~Su, ``Tensorf: Tensorial radiance fields,'' in \emph{European Conference on Computer Vision}.\hskip 1em plus 0.5em minus 0.4em\relax Springer, 2022, pp. 333--350.

\bibitem{sun2021direct}
C.~Sun, M.~Sun, and H.-T. Chen, ``Direct voxel grid optimization: Super-fast convergence for radiance fields reconstruction,'' \emph{CVPR}, 2022.

\bibitem{turki2022mega}
H.~Turki, D.~Ramanan, and M.~Satyanarayanan, ``Mega-nerf: Scalable construction of large-scale nerfs for virtual fly-throughs,'' in \emph{Proceedings of the IEEE/CVF Conference on Computer Vision and Pattern Recognition}, 2022, pp. 12\,922--12\,931.

\bibitem{tancik2022block}
M.~Tancik, V.~Casser, X.~Yan, S.~Pradhan, B.~Mildenhall, P.~P. Srinivasan, J.~T. Barron, and H.~Kretzschmar, ``Block-nerf: Scalable large scene neural view synthesis,'' in \emph{Proceedings of the IEEE/CVF Conference on Computer Vision and Pattern Recognition}, 2022, pp. 8248--8258.

\bibitem{mi2023switchnerf}
\BIBentryALTinterwordspacing
Z.~Mi and D.~Xu, ``Switch-nerf: Learning scene decomposition with mixture of experts for large-scale neural radiance fields,'' in \emph{International Conference on Learning Representations (ICLR)}, 2023. [Online]. Available: \url{https://openreview.net/forum?id=PQ2zoIZqvm}
\BIBentrySTDinterwordspacing

\bibitem{deng2022depth}
K.~Deng, A.~Liu, J.-Y. Zhu, and D.~Ramanan, ``Depth-supervised nerf: Fewer views and faster training for free,'' in \emph{Proceedings of the IEEE/CVF Conference on Computer Vision and Pattern Recognition}, 2022, pp. 12\,882--12\,891.

\bibitem{bian2023nope}
W.~Bian, Z.~Wang, K.~Li, J.-W. Bian, and V.~A. Prisacariu, ``Nope-nerf: Optimising neural radiance field with no pose prior,'' in \emph{Proceedings of the IEEE/CVF Conference on Computer Vision and Pattern Recognition}, 2023, pp. 4160--4169.

\bibitem{sandstrom2023point}
E.~Sandstr{\"o}m, Y.~Li, L.~Van~Gool, and M.~R. Oswald, ``Point-slam: Dense neural point cloud-based slam,'' in \emph{Proceedings of the IEEE/CVF International Conference on Computer Vision}, 2023, pp. 18\,433--18\,444.

\bibitem{zhu2022nice}
Z.~Zhu, S.~Peng, V.~Larsson, W.~Xu, H.~Bao, Z.~Cui, M.~R. Oswald, and M.~Pollefeys, ``Nice-slam: Neural implicit scalable encoding for slam,'' in \emph{Proceedings of the IEEE/CVF Conference on Computer Vision and Pattern Recognition}, 2022, pp. 12\,786--12\,796.

\bibitem{rosinol2023nerf}
A.~Rosinol, J.~J. Leonard, and L.~Carlone, ``Nerf-slam: Real-time dense monocular slam with neural radiance fields,'' in \emph{2023 IEEE/RSJ International Conference on Intelligent Robots and Systems (IROS)}.\hskip 1em plus 0.5em minus 0.4em\relax IEEE, 2023, pp. 3437--3444.

\bibitem{sucar2021imap}
E.~Sucar, S.~Liu, J.~Ortiz, and A.~J. Davison, ``imap: Implicit mapping and positioning in real-time,'' in \emph{Proceedings of the IEEE/CVF International Conference on Computer Vision}, 2021, pp. 6229--6238.

\bibitem{yuan2022nerf}
Y.-J. Yuan, Y.-T. Sun, Y.-K. Lai, Y.~Ma, R.~Jia, and L.~Gao, ``Nerf-editing: geometry editing of neural radiance fields,'' in \emph{Proceedings of the IEEE/CVF Conference on Computer Vision and Pattern Recognition}, 2022, pp. 18\,353--18\,364.

\bibitem{mirzaei2023spin}
A.~Mirzaei, T.~Aumentado-Armstrong, K.~G. Derpanis, J.~Kelly, M.~A. Brubaker, I.~Gilitschenski, and A.~Levinshtein, ``Spin-nerf: Multiview segmentation and perceptual inpainting with neural radiance fields,'' in \emph{Proceedings of the IEEE/CVF Conference on Computer Vision and Pattern Recognition}, 2023, pp. 20\,669--20\,679.

\bibitem{Kong2023RoomNeRF}
\BIBentryALTinterwordspacing
M.~Kong, S.~Lee, and E.~Kim, ``Roomnerf: Representing empty room as neural radiance fields for view synthesis,'' in \emph{British Machine Vision Conference}, 2023. [Online]. Available: \url{https://api.semanticscholar.org/CorpusID:267000673}
\BIBentrySTDinterwordspacing

\bibitem{pumarola2021d}
A.~Pumarola, E.~Corona, G.~Pons-Moll, and F.~Moreno-Noguer, ``D-nerf: Neural radiance fields for dynamic scenes,'' in \emph{Proceedings of the IEEE/CVF Conference on Computer Vision and Pattern Recognition}, 2021, pp. 10\,318--10\,327.

\bibitem{fridovich2023k}
S.~Fridovich-Keil, G.~Meanti, F.~R. Warburg, B.~Recht, and A.~Kanazawa, ``K-planes: Explicit radiance fields in space, time, and appearance,'' in \emph{Proceedings of the IEEE/CVF Conference on Computer Vision and Pattern Recognition}, 2023, pp. 12\,479--12\,488.

\bibitem{li2023dynibar}
Z.~Li, Q.~Wang, F.~Cole, R.~Tucker, and N.~Snavely, ``Dynibar: Neural dynamic image-based rendering,'' in \emph{Proceedings of the IEEE/CVF Conference on Computer Vision and Pattern Recognition}, 2023, pp. 4273--4284.

\bibitem{zhi2021place}
S.~Zhi, T.~Laidlow, S.~Leutenegger, and A.~J. Davison, ``In-place scene labelling and understanding with implicit scene representation,'' in \emph{Proceedings of the IEEE/CVF International Conference on Computer Vision}, 2021, pp. 15\,838--15\,847.

\bibitem{cen2023segment}
J.~Cen, Z.~Zhou, J.~Fang, W.~Shen, L.~Xie, D.~Jiang, X.~Zhang, Q.~Tian \emph{et~al.}, ``Segment anything in 3d with nerfs,'' \emph{Advances in Neural Information Processing Systems}, vol.~36, pp. 25\,971--25\,990, 2023.

\bibitem{kendall2015posenet}
A.~Kendall, M.~Grimes, and R.~Cipolla, ``Posenet: A convolutional network for real-time 6-dof camera relocalization,'' in \emph{Proceedings of the IEEE international conference on computer vision}, 2015, pp. 2938--2946.

\bibitem{mildenhall2019local}
B.~Mildenhall, P.~P. Srinivasan, R.~Ortiz-Cayon, N.~K. Kalantari, R.~Ramamoorthi, R.~Ng, and A.~Kar, ``Local light field fusion: Practical view synthesis with prescriptive sampling guidelines,'' \emph{ACM Transactions on Graphics (TOG)}, vol.~38, no.~4, pp. 1--14, 2019.

\bibitem{lin2020nerfpytorch}
L.~Yen-Chen, ``Nerf-pytorch,'' \url{https://github.com/yenchenlin/nerf-pytorch/}, 2020.

\bibitem{hedman2018deep}
P.~Hedman, J.~Philip, T.~Price, J.-M. Frahm, G.~Drettakis, and G.~Brostow, ``Deep blending for free-viewpoint image-based rendering,'' \emph{ACM Transactions on Graphics (ToG)}, vol.~37, no.~6, pp. 1--15, 2018.

\end{thebibliography}

\end{document}